\documentclass[lettersize,journal]{IEEEtran}
\usepackage{amsmath,amsfonts}
\usepackage{algorithmic}
\usepackage{algorithm}
\usepackage{array}
\usepackage[caption=false,font=normalsize,labelfont=sf,textfont=sf]{subfig}
\usepackage{textcomp}
\usepackage{stfloats}
\usepackage{url}
\usepackage{verbatim}
\usepackage{graphicx}
\usepackage{cite}

\usepackage{times}
\usepackage{epsfig}
\usepackage{amssymb}
\usepackage{booktabs}
\usepackage{multirow}
\usepackage{subcaption}
\usepackage{makecell}
\usepackage{arydshln}
\usepackage{pifont}
\usepackage[pagebackref,breaklinks,colorlinks]{hyperref}
\usepackage[capitalize]{cleveref}
\crefname{section}{Sec.}{Secs.}
\Crefname{section}{Section}{Sections}
\Crefname{table}{Table}{Tables}
\crefname{table}{Tab.}{Tabs.}
\newcommand{\etal}{\textit{et al}. }
\newcommand{\ie}{\textit{i}.\textit{e}.,}
\newcommand{\eg}{\textit{e}.\textit{g}.}

\title{Revealing Directions for Text-guided \\3D Face Editing}

\author{Zhuo Chen,
Yichao Yan,
Sehngqi Liu, 
Yuhao Cheng, \\
Weiming Zhao, 
Lincheng Li,
Mengxiao Bi,
Xiaokang Yang \\ 
\thanks{(Corresponding author: Yichao Yan.)}
\thanks{Zhuo Chen, Yichao Yan, Sehngqi Liu, Yuhao Cheng and Xiaokang Yang are with the MoE Key Lab of Artificial Intelligence, AI Institute, Shanghai Jiao Tong University, Shanghai, China. (email: ningci5252@sjtu.edu.cn, yanyichao@sjtu.edu.cn, lsqlsq@sjtu.edu.cn, chengyuhao@sjtu.edu.cn, xkyang@sjtu.edu.cn)}
\thanks{Weiming Zhao is with the Student Innovation Center, Shanghai Jiao Tong University, Shanghai, China. (email: weiming.zhao@sjtu.edu.cn)}
\thanks{Lincheng Li and Mengxiao Bi are with NetEase Fuxi AI Lab, Hangzhou, China. (email: lilincheng@corp.netease.com, bimengxiao@corp.netease.com)}
}

\begin{document}
\let\oldtwocolumn\twocolumn
\renewcommand\twocolumn[1][]{%
	\oldtwocolumn[{#1}{
		\begin{center}
			\includegraphics[width=\linewidth]{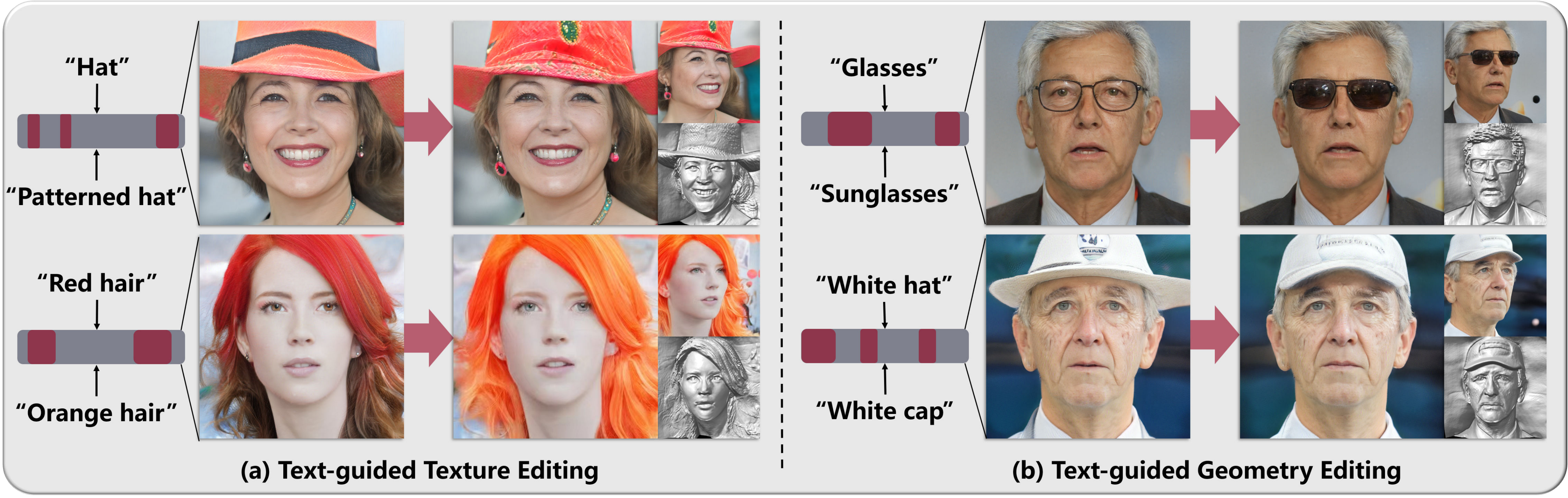}
			\captionof{figure}{Examples of our text-guided 3D face editing. Our method links the input text to a mask on latent codes, enabling attribute disentanglement and identity preservation. It has the capacity of both (a) texture editing and (b) geometry manipulation.} 
			\label{fig:teaser}
		\end{center}
	}]
}

\maketitle

\begin{abstract}
3D face editing is a significant task in multimedia, aimed at the manipulation of 3D face models across various control signals.
The success of 3D-aware GAN provides expressive 3D models learned from 2D single-view images only, encouraging researchers to discover semantic editing directions in its latent space.
However, previous methods face challenges in balancing quality, efficiency, and generalization.
To solve the problem, we explore the possibility of introducing the strength of diffusion model into 3D-aware GANs.
In this paper, we present \textbf{Face Clan}, a fast and text-general approach for generating and manipulating 3D faces based on arbitrary attribute descriptions. 
To achieve disentangled editing, we propose to diffuse on the latent space under a pair of opposite prompts to estimate the mask indicating the region of interest on latent codes. 
Based on the mask, we then apply denoising to the masked latent codes to reveal the editing direction.
Our method offers a precisely controllable manipulation method, allowing users to intuitively customize regions of interest with the text description. Experiments demonstrate the effectiveness and generalization of our Face Clan for various pre-trained GANs. It offers an intuitive and wide application for text-guided face editing that contributes to the landscape of multimedia content creation.
\end{abstract}



\section{Introduction}\label{sec:intro}

3D face editing is an essential task in multimedia, aiming to manipulate face attributes in a 3D representation under various control signals, while preserving the original identity and 3D consistency.
In 2D editing, we have witnessed the creative and diverse synthesis achieved by diffusion models. However, it is hard to directly lift this talent to 3D generation due to its heavy dependency on the 3D ground-truth data.
Fortunately, recent 3D-aware GANs~\cite{schwarz2020graf, chan2020pi, Niemeyer2020GIRAFFE, gu2021stylenerf,zhou2021cips,or2022stylesdf,chan2022efficient,xu20223d,deng2022gram, xiang2023gram,skorokhodov2022epigraf,wang2023rodin,an2023panohead,yan_survey} brings the ability of synthesizing natural 3D representation by learning from unposed single-view 2D images only. 
Moreover, a GAN model~\cite{karras2019style,karras2020analyzing} learns well-disentangled attributes within its latent space and supports an arbitrary combination of attributes, which is equivalent to compacting the composite 3D face into a light latent code.
The light representation and rich distributions encourage researchers to explore the meaningful directions in the latent space, and also potentially empower us to collaborate with the power of mature editing strategy based on the diffusion model.


Previous methods have explored semantic directions in the latent space, but they still face challenges in balancing intuitiveness, generalization, and efficiency.
Supervised methods~\cite{shen2020interfacegan,abdal2021styleflow,patashnik2021styleclip,simsar2023latentswap3d, 10100897} 
are time-consuming and incapable of the attribute whose classifiers are not available, while unsupervised methods~\cite{voynov2020unsupervised,harkonen2020ganspace,shen2021closed,zhu2021low,zhu2022region,zhu2023linkgan} are highly sensitive to the analyzed identity and fail to discover arbitrary semantic directions as desired by the user.
Another category of work introduces additional control signals to discover the editing direction.
These methods optimize or predict latent codes under the guidance of various conditions, \eg, semantic map~\cite{sun2022ide,jiang2022nerffaceediting}, 3DMM~\cite{tewari2020stylerig,sun2023next3d} and dragging points~\cite{endo2022user,pan2023drag, Cheng_2024_CVPR}. 
Although these conditioned methods have achieved success in shape and expression manipulation, they encounter difficulties when attempting to edit color and texture.


To achieve a more general and intuitive way to manipulate 3D faces, there has been a growing interest on text-guided exploration of semantic directions.
Recent text-conditioned methods can be roughly classified into 1) optimization in latent space or parameter space and 2) an encoder that directly projects CLIP features to latent space.
Optimization methods~\cite{gal2022stylegan, chen2023hyperstyle3d} guided by CLIP~\cite{radford2021learning} or Stable diffusion~\cite{rombach2022high} can handle general texts, albeit at the cost of prolonged editing times.
To improve efficiency, recent works~\cite{li20243d, zhang2024fast, yu2023towards, zhang2023styleavatar3d, 9737433} align the text feature with latent codes via an encoder. However, the one-step determined projection reduces their capacity and leads to a lack of diversity and identity preservation.
Thus, there is an inherent need for \textbf{an optimization-free model that can balance text generalization, editing quality, and efficiency}.

Inspired by the success of diffusion models in text-to-2D synthesis~\cite{meng2021sdedit, couairon2022diffedit, avrahami2022blended, lugmayr2022repaint,hertz2022prompt, kawar2023imagic, rombach2022high}, we come out with an idea that \textbf{diffuses the latent space of GAN} to align text conditions with latent codes instead of 2D images.
In contrast to the mapper that directly projects text features into latent spaces, the diffusion model in this work demonstrates superior text-to-latent diversity and consistency, due to its multi-step accumulated bias as a generative model~\cite{ho2020denoising,song2020denoising, nichol2021improved}.
Text conditions can be regarded as a direction indicator that leads the sampled noise to the class consistent with the given text. This special characteristic can both enhance text-guided generation quality and facilitate editing direction reveal.
Based on the inspiration, we present a fast and text-generalized approach called \textbf{Face Clan}, to automatically generate and manipulate the 3D face based on arbitrary attribute descriptions by reconstructing the distribution of latent codes.

A robust method with high quality should take effect on the contents within the area of interest only and preserve the remaining contents as much as possible. We decompose it into \textbf{mask} and \textbf{direction} in the latent space.
\textbf{1) To edit the contents}, we need to design a text-guided model that reveals an editing \textbf{direction} for latent codes.
\textbf{2) To preserve the remaining parts}, 
a direct idea is to apply a mask on the region of interest. However, latent space is visually implicit so that users cannot intuitively add the mask. 
Therefore, it is supposed to analyze the interest region of given texts and link it to a \textbf{mask} on latent codes.

Face Clan first trains a diffusion model to align the distribution of the text with the learned latent manifolds of pre-trained 3D-aware GAN in a self-supervised way.
Given a set of synthesized data from the pre-trained GAN, it is capable of learning to map a randomly sampled Gaussian noise to a text-consistent latent code with identity diversity.
Modulated by the latent code, the 3D-aware generator can further produce a 3D-consistent face according to the textual description.
Based on the trained text-to-3D-face model, we are empowered to precisely edit face attributes.
During editing, we initially estimate a text-relevant mask on the latent code by measuring the principal difference between two predicted noises under a pair of opposite descriptions, \eg, \textit{``hat''} and \textit{``cap''} as despite in \cref{fig:teaser}.
With the mask, a denosing procedure can be performed on the masked region of noisy latent codes, while keeping the unmasked region as usual.
Consequently, it is flexible that users can customize the desired regions under input prompts. 
Extensive experimental results show that our Face Clan achieves precise controllability and strong robustness in various 3D-aware GANs.
The main contributions are summarized as follows: 
\begin{itemize}
\setlength\itemsep{0em}
\item We design a fast and generalized text-guided face editing pipeline based on a self-supervised diffusion model that aligns distributions of texts and latent manifolds in the pre-trained GAN.
\item We propose a directional mask estimation to link the text to the region of interest in latent codes, achieving robust attribute disentanglement.
\item Our editing approach can handle identity-specific attributes out of the common face attribute domain, \eg, hat style as despite in \cref{fig:teaser}.
\end{itemize}

\section{Related Works}\label{sec:2_related_works}

\subsection{Semantic Direction Exploration in GANs}
\label{sub21}
The latent space of GANs contains a wealth of semantic features, facilitating the exploration of diverse semantic directions for image editing. 
Face editing methods in the latent space can be broadly categorized as \textbf{condition-guided direction discovery} and \textbf{unconditioned direction discovery}
Unconditioned direction discovery can be further subdivided into supervised and unsupervised exploration. 
Supervised methods~\cite{shen2020interfacegan,abdal2021styleflow,patashnik2021styleclip,simsar2023latentswap3d} involve labeling image data with the target binary attribute and training a classifier to reveal the inherent hyperplane.
The normal of this hyperplane indicates the editing direction between the binary attribute.
While this approach is stable and effective, it is laborious to discover an editing direction due to the need for labeled data and additional classifier training for each attribute.
Some unsupervised methods~\cite{voynov2020unsupervised,harkonen2020ganspace,shen2021closed,zhu2021low,zhu2022region,zhu2023linkgan, 9380493} analyze the principle feature of the latent space to discover meaningful editing directions. 
However, these methods lack the intuitiveness of semantic directions, as these directions in editing images often need to be manually discerned. Besides, it is hard to generalize to diverse attributes, limited to the principal face attributes.
To achieve more intuitive editing, some methods explore semantic directions through explicit conditions, \eg, 3DMM~\cite{tewari2020stylerig,sun2023next3d}, semantic maps~\cite{sun2022ide,jiang2022nerffaceediting}, and point dragging~\cite{endo2022user,pan2023drag, Cheng_2024_CVPR}. 
These methods incorporate multi-modality into facial editing of multimedia, \eg, image, video, and 3D representations.

Specifically, Zhu \etal~\cite{10444967} separate the face into the overall shape and detailed regions, utilizing 3DMM to guide shape deformation and semantic mask to guide region manipulation. Yang \etal~\cite{10246440} also take region masks to edit the latent code for facial editing. Unlike previous works, GlassesCLIP~\cite{10274147} focuses on the specific glasses region, which is challenging in facial editing.
Despite their effectiveness and intuitiveness, these methods are constrained to geometry manipulation due to the representative ability of these conditions. Failure in texture editing restricts their applications.
Among the vast control signals, natural language is the primary means of human expression, containing both geometry and texture semantics that can align with image attributes and realize more diverse editing on targets.

\begin{figure*}[t]
	\centering
	\includegraphics[width=\linewidth]{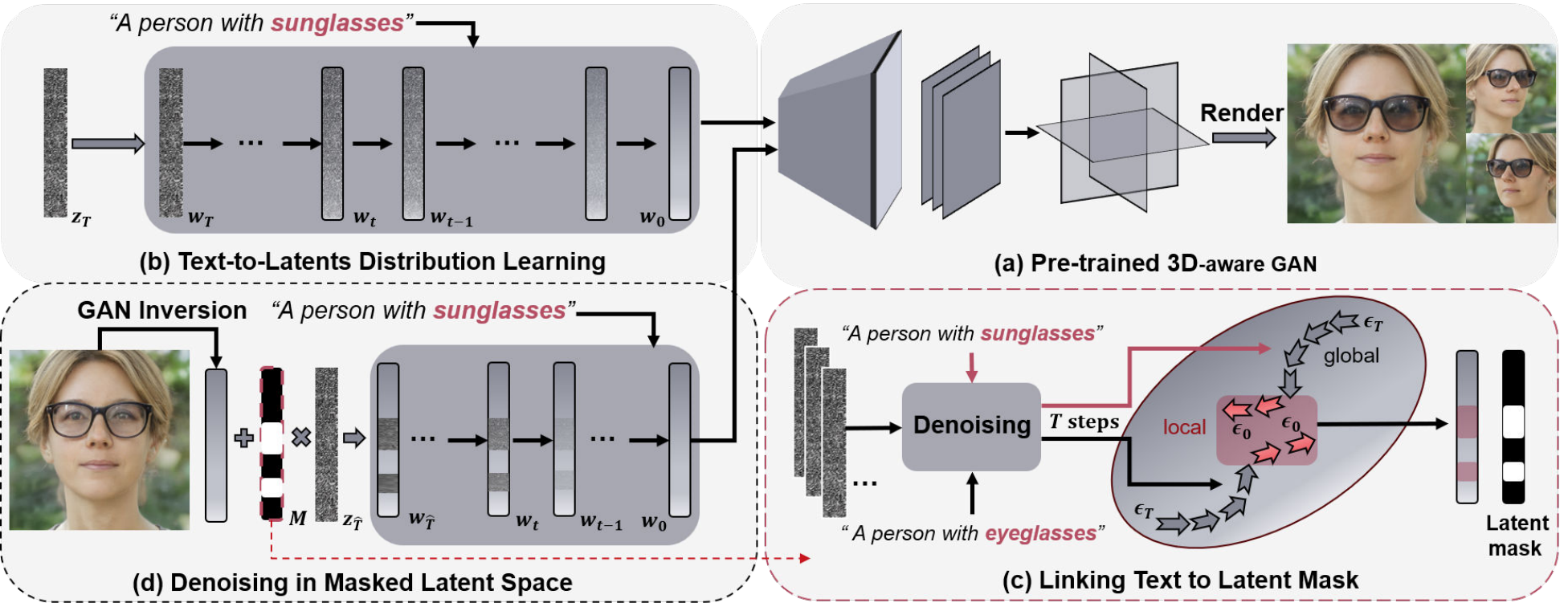}
	\caption{Overview of our proposed method. (a) The architecture of our based 3D-aware generator, EG3D. (b) The inference pipeline to redistribute the latent code for text-guided synthesis. (c) The illustration of linking text to the region of interest on latent codes. (d) Apply denoising to masked latent codes for disentangled face editing.
    }
	\label{img:overview}
\end{figure*}

\subsection{Text-guided Editing in GANs}
In the realm of multimedia innovation, some researchers delve into the intricate interplay between textual descriptions and image manipulation, offering an intuitive approach to face editing.
Some earlier approaches~\cite{dong2017semantic,nam2018text,liu2020describe,li2020manigan} utilize encoder-decoder structures to construct the relationship between images and text for editing. 
To enable semantic editing, some approaches~\cite{patashnik2021styleclip,gal2022stylegan,zhang2024fast,li20243d,yu2023towards} propose editing within GANs, leveraging pre-trained large-scale multi-modal models, \eg, CLIP~\cite{radford2021learning}.
StyleCLIP~\cite{patashnik2021styleclip} introduces a paradigm to optimize the latent code of GANs. StyleGAN-Nada~\cite{gal2022stylegan} proposes to fine-tune the generator of GANs to improve a specific type of editing. 
MorphNeRF~\cite{10476703} proposes a learnable network that morphs the 3D geometry of faces toward the text descriptions via NeRF-based GAN and CLIP.
Despite high-quality editing results, they suffer from inefficiency due to the time-consuming optimization.
Aiming at efficiency improvement, the following works~\cite{zhang2024fast,li20243d,yu2023towards, du2023pixelface+, peng2022learning, peng2022towards} directly project the text feature extracted by the CLIP model to the latent space. 
TextFace~\cite{9737433} introduces text-to-style mapping 
based on text-image similarity matching and a face-caption alignment, achieving high-fidelity face generation and manipulation.
However, the one-step determined projection limits their capacity, leading to conflict between diversity and identity preservation.
Therefore, it is essential to propose a method that can balance the quality and efficiency of editing.
Our proposed Face Clan leverages a lightweight diffusion model to map texts to latent codes without optimizing the generators, achieving efficient and wide-ranging face editing.

\subsection{Diffuse the Latent Space of GANs}
The success of diffusion in 2D image editing has been widely witnessed.
These methods either focus on textural inversion~\cite{hertz2022prompt, kawar2023imagic} or redistribute the image via noising and denoising~\cite{meng2021sdedit, couairon2022diffedit, avrahami2022blended, lugmayr2022repaint}. 
Specifically, MMGInpainting~\cite{10480591} incorporates an Anchored Stripe Attention mechanism that utilizes anchor points to model global contextual dependencies, effectively integrating the semantic information into the target region of faces. 
We draw inspiration from these methods, \eg, SDEdit~\cite{meng2021sdedit} that adds a controlled level of noise to edit towards the target while preserving the partial structure of the original image.
An idea commonly employed for precise editing, as demonstrated by Repaint~\cite{lugmayr2022repaint}, Blend Diffusion~\cite{avrahami2022blended}, and DiffEdit~\cite{couairon2022diffedit}, involves the introduction of masks.
In contrast, our focus lies in the latent space of GANs, where the latent codes, unlike 2D images, are well-distributed and lightweight. This characteristic means that the editing direction of an attribute is nearly unified across all codes, and revealing the direction from a group of codes is fast and efficient.
The advantages of the latent code and the diffusion model have motivated researchers to explore the possibility of their combination.
Recent works~\cite{wang2023rodin, shen2024controllable, zhang2023styleavatar3d, lei2023diffusiongan3d, kirschstein2023diffusionavatars, li20233d} connect a diffusion model to the latent space of GANs, enabling text-controllable synthesis. However, these approaches have not addressed the preservation of identities, thereby failing in precise editing. The work most closely related to ours is InstructPix2NeRF~\cite{li2023instructpix2nerf} (ICLR 2024), which employs a conditional 3D diffusion to lift 2D editing to 3D space through the acquisition of the correlation between the image and the instruction.
Despite diverse and generalized editing, it still falls short in disentanglement and identity preservation. 
To enhance the editing precision, 
we design a method for mask estimation on latent codes to realize more controllable editing.


\section{Methods}
In this section, 
we begin by reviewing the 3D-aware GAN and the diffusion model that respectively produces high-quality 3D faces and learns text-conditioned latent distribution (\cref{sec:preliminaries}). 
Benefiting from the rich semantics learned in the latent space of GANs, we design a self-supervised diffusion model that aligns the distributions of text prompts and latent codes, to achieve text-guided face synthesis. (\cref{sec:swap}).
Based on this trained text-to-3D-face model, we further propose a text-guided editing pipeline, which consists of two specific steps, \ie mask estimation (\cref{sec:projection}) and target direction discovery (\cref{sec:edit}) on latent codes. The overview of our method is shown in \cref{img:overview}.

\subsection{Preliminaries}
\label{sec:preliminaries}
\noindent\textbf{3D-aware GAN.}
Considering the generation quality, our framework is constructed based on EG3D~\cite{chan2022efficient}, as shown in \cref{img:overview}~(a).
Similar to other conditioned GANs, EG3D contains a mapping network $f(\cdot)$ responsible for projecting random latent code $z$ (together with camera parameters) to an intermediate latent code $\mathbf{w}=f(z)\in \mathbb{R}{^{512}}$, which subsequently modulates a synthesis network.
This latent code $\mathbf{w}$ plays a crucial role in face editing, as it learns disentangled attributes and aligns with the data distribution.
Conditioned on $\mathbf{w}$, the generator of EG3D further produces a tri-plane feature $\mathbf{F} \in \mathbb{R}^{3 \times 32 \times 256 \times 256}$ as a 3D representation. 
Subsequently, a shallow MLP decoder projects the tri-plane feature $\mathbf{F}$ into volume density \textbf{$\sigma$} and color feature \textbf{$c$}, which are further rendered into high-resolution images via volume rendering and a super-resolution module. 
The entire pipeline can be simplified as
$I_{0} = \mathcal{G}\left(\mathbf{w_0} \right) = \mathcal{G}\left(f(z_0) \right)$, where $\mathcal{G}$ represents the generator of EG3D.
Compared with the inherent 3D representation $\mathbf{F}\in\mathbb{R}^{3 \times 32 \times 256 \times 256}$, a latent code $\mathbf{w}\in \mathbb{R}{^{512}}$ is heavily compacted, while it still contains most information of a 3D face. Moving along a semantic direction within the latent space allows for easy face editing, 
$$I'_{0} = \mathcal{G}\left(\mathbf{w_0} + \Delta \mathbf{w}\right).$$

\noindent\textbf{Latent Diffusion Model.}
The diffusion model has shown its strength in diverse generations.
It incorporates a forward process, given by the equation:
$$q(\mathbf{x}_{1:T}|\mathbf{x}_0) = \prod^T_{t=1} q(\mathbf{x}_t \vert \mathbf{x}_{t-1}),$$
which add noise to the latent representation of the source image $\mathbf{x}_0$, and a reverse process, given by the equation:
$$p(\mathbf{x}_{0:T}) = p(\mathbf{x}_T) \prod^T_{t=1} p_\theta(\mathbf{x}_{t-1} \vert \mathbf{x}_t),$$
which utilizes a parameterized denoising network $\theta$ to gradually denoise the target latent representation.
To ensure that the forward process is approximately equal to the reverse process, the latent diffusion model is trained by minimizing weighted evidence lower bound:
\begin{equation}
    \mathcal{L}_{\mathrm{Diff}} =\mathbb{E}_{t,\epsilon}\left[w(t)\|\epsilon_{\theta}(\mathbf{x}_{t};t)-\epsilon\|_{2}^{2}\right],
\end{equation}
where $w(t)$ is a weighting function, time step $t\sim\mathcal{U}(0,1)$, random noise $\epsilon\sim\mathcal{N}(\mathbf{0},\mathbf{I})$, and $\theta$ denotes the parameters of the denoising network.
After the training, we can randomly sample a Gaussian noise input $\mathbf{x}_T \sim \mathcal{N}(\mathbf{0}, \mathbf{I})$ and apply denoising on it to generate the target $x_{0}$ conforming to data distribution.

\subsection{Text-to-Latents Distribution Learning}
\label{sec:swap}
The diffusion model possesses a strong capability to project Gaussian noise onto a sample from the data distribution, making it well-suited for the mapping network $f(z)$ in GANs. Consequently, we introduce a diffusion model as an alternative mapping network into the latent space of GAN, to learn the mapping from $Z$ space to $W$ space, as shown in \cref{img:overview}~(b).
Following the preliminaries, we train a diffusion model whose $\theta$ is parameterized by DiT~\cite{yang2023diffusion} for the latent space of GANs.
To further enable explicit-conditioned generation based on a pre-trained generator, we add a cross-attention layer~\cite{chen2023pixart} to introduce conditions $y$ and train the model by an objective~\cite{rombach2022high},
\begin{equation}
L_{LDM}=\mathbb{E}_{\epsilon \sim \mathcal{N}(0,1), y, t}\left[\left\|\epsilon-\epsilon_\theta\left(w_t, t, \tau_\theta(y)\right)\right\|_2^2\right].
\end{equation}
Although training a latent diffusion model typically raises concerns about the requirement for a large amount of paired image-text data, we can solve it through self-supervised data generation.
Following previous works~\cite{gu2023learning}, we can synthesize pairs of the condition and the latent code infinitely by the synthesized images from the 3D generator and a feature extractor for the target condition.
While our focus in this work is on the text condition, it is worth noting that automatic caption models~\cite{li2023blip} exhibit poor performance and limited diversity when applied to human faces.
Fortunately, the aligned space of CLIP~\cite{radford2021learning} allows us to utilize CLIP image embedding as the condition during training, while using text embedding during inference, as in previous methods~\cite{gu2023learning, li20233d, pinkney2022clip2latent}.
With the CLIP feature extractor $C(\cdot)$, the training objective can be written as,
\begin{equation}
L_{LDM}=\mathbb{E}_{\epsilon \sim \mathcal{N}(0,1), C\left(\mathcal{G}\left(w_0 \right)\right), t}\left[\left\|\epsilon-\epsilon_\theta\left(w_t, t, C\left(\mathcal{G}\left(w_0 \right)\right)\right)\right\|_2^2\right].
\end{equation}
To improve training efficiency, both data generation and feature extraction are off-the-shelf.

\subsection{Linking Text to Latent Mask}
\label{sec:projection}
\begin{figure}[t]
	\centering
	\includegraphics[width=1.0\linewidth]{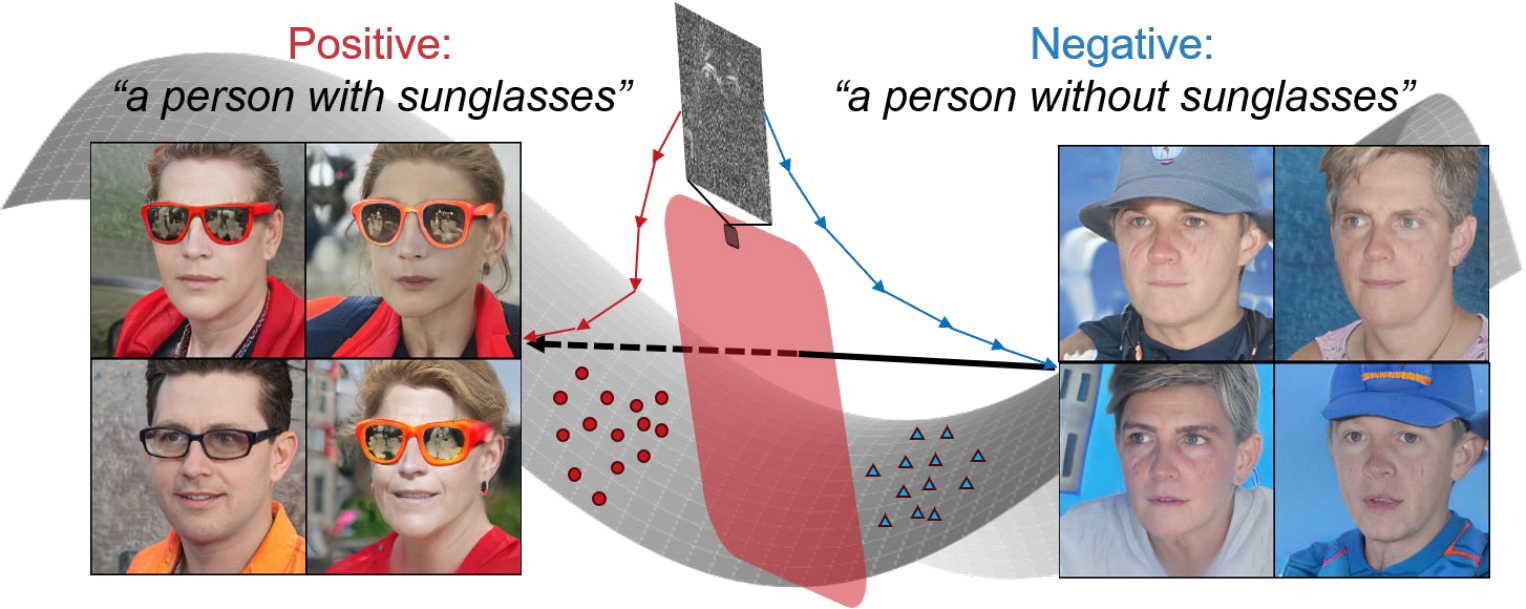}
	\caption{The illustration of the inspiration that adopts diffusion model as both a controllable data generator and an attribute classifier.}
	\label{img:relative}
\end{figure}

Common 2D image editing encourages applying a mask to keep irrelevant pixels the same within the region of interest.
Different from direct 2D image editing, there is no visually explicit semantics in latent codes, which makes it hard to directly apply masks as users' intuitiveness.
To intuitively control local semantics in the output 3D face, it leads to a demand that links the explicit meaningful text to a mask on the implicit latent code.
Inspired by the supervised methods that collect labeled samples with binary attributes to classify, we come out with an idea that we can treat our diffusion model as a natural classifier to both generate samples and find directions between them, as shown in \cref{img:relative} and \cref{img:overview}~(a).

Specifically, to reveal a collective direction of a specific attribute, we first sample several Gaussian noises $\mathbf{Z}_{T}=\{\mathbf{z}_{T}^0, \mathbf{z_{T}}^1, \cdots, \mathbf{z}_{T}^n\}$ and further denoise under the condition of paired descriptions with opposite attributes $y_{\rm src}$ and $y_{\rm tgt}$, \eg, \textit{``a person with sunglasses''} and \textit{``a person without sunglasses''}. The detailed procedure of each step is represented as, 
\begin{equation}
\mathbf{Z}_{t-1}=\sqrt{\alpha_{t-1}}\left(\frac{\mathbf{Z}_t-\sqrt{1-\alpha_t} \cdot \epsilon_t}{\sqrt{\alpha_t}}\right)+\sqrt{1-\alpha_{t-1}} \cdot \epsilon_t ,
\end{equation}
\begin{equation}
 \epsilon_t = \epsilon_\theta\left(\mathbf{Z}_t, t, \tau_\theta\left( y\right)\right),
\end{equation}
where noise schedule $\alpha_t \in (0, 1)$, $\sqrt{1-\alpha_{t-1}} \cdot \epsilon_t$ indicates the direction to the target $Z_0$.
To simplify the representation, we donate the full denoising procedure $D$ as, 
\begin{equation}
\mathbf{W}_{\rm src} = \mathbf{Z}_{0}^{src} = D\left(\mathbf{Z}_{T}, y_{src} \right),
\end{equation}
\begin{equation}
\mathbf{W}_{\rm tgt} = \mathbf{Z}_{0}^{tgt} = D\left(\mathbf{Z}_{T}, y_{tgt} \right),
\end{equation}
where $\mathbf{W}_{\rm src}$ and $\mathbf{W}_{\rm tgt}$ are the denoised latent codes of the pre-trained model under opposite prompts $y_{src}$ and $y_{tgt}$.

\textbf{Mask estimated from $\mathbf{\epsilon}$ or $\mathbf{W}$.}
With two groups of latent codes, a naive idea is to estimate the mask $M$ by the difference as, 
\begin{equation}
M[i] = 
\begin{cases}
1, Norm\left(\left|\mathbf{W}_{tgt} - \mathbf{W}_{src} \right|\right)[i] > threshold, \\
0, otherwise,
\end{cases}
\end{equation}
where $i$ is the $i$th dimension in the latent code.
It is effective but suffers from attribute entanglement. The reason is that the difference between paired latent codes can be seen as accumulating the bias of full-step paired noises. However, similar to 2D image diffusion models, predicted noises in different steps indicate directions for different-level attributes~\cite{wang2023diffusion,huang2023reversion,han2023svdiff}.
The endpoint estimate $\hat{x_0}$ travels on the latent manifold, initiating from the center of the distribution, moving first along the high variance axes, and then the lower variance axes.
Consequently, low-level attributes are produced in the early stage, while finer details tend to emerge in the last few steps.
Therefore, the noise difference accumulated in latent codes contains more unexpected global changes besides the local target attribute.
Fig.~\ref{img:ablation} in our ablation study also supports this theory.
According to the above analysis, we turn to estimate the mask by the difference between paired predicted noises in the last few steps. It avoids the influence of global and fused changes led by the early stage. The procedure is represented as, 
\begin{equation}
 \epsilon_{t}^{\rm src} = \epsilon_\theta\left(\mathbf{Z}_t, t, \tau_\theta\left( y_{\rm src}\right)\right),
\end{equation}
\begin{equation}
 \epsilon_{t}^{\rm tgt} = \epsilon_\theta\left(\mathbf{Z}_t, t, \tau_\theta\left( y_{\rm tgt}\right)\right),
\end{equation}
\begin{equation}
M[i] = 
\begin{cases}
1, Norm\left(\left|\epsilon_{t}^{\rm tgt} - \epsilon_{t}^{\rm src} \right|\right)[i] > threshold, \\
0, otherwise.
\end{cases}
\end{equation}

With the estimated mask from noise direction, a simple idea for face editing is to directly swap the $\mathbf{W}_{\rm tgt}$ with $\mathbf{W}_{\rm src}$ in the masked region as the editing direction, which is represented as, 
\begin{equation}
\Delta \mathbf{w} = M \odot \left(\mathbf{\bar W}_{\rm tgt} - \mathbf{\bar W}_{\rm src} \right),
\end{equation}
\begin{equation}
I_{\rm edit} = \mathcal{G}\left(\mathbf{w}_{\rm input} + \alpha \Delta \mathbf{w}\right).
\end{equation}
Unfortunately, a precise latent mask can constrain the region dimensions in the latent code, but can not accurately indicate the change direction. It neither achieves a large-scale change for the target attribute nor preserves identity during editing towards this direction.
Despite a scalable $\alpha$ to manipulate the aptitude, a large scale $\alpha$ with the misdirection $\Delta \mathbf{w}$ leads to corner samples in the distribution. 
We have conducted an ablation study in \cref{img:ablation} to support the claim. Therefore, we need an additional step to ascertain the efficient direction in the masked latent space.

\subsection{Denoising in Masked Latent Space}
\label{sec:edit}

Given an input image $\mathbf{I}_e$, we first invert it into a latent code $\mathbf{w}_e$,
\begin{align}
    \mathbf{z}_{0}^e = \mathbf{w}_e = \mathcal{E}(\mathbf{I}_e).
    \label{encoder_align}
\end{align}
With the estimated mask $M$, we can perform text-conditioned diffusion on the masked region of the latent code $\mathbf{z}_{0}^e$ to reveal the efficient direction.
Compared to 2D image editing, latent codes are much less sensitive on the edge between masked and unmasked regions, obviating concerns regarding edge inconsistency.
Specifically, we add noise to the latent code $\mathbf{z}_{0}^e$ instead of sampling several Gaussian noises,
\begin{equation}
q\left(\mathbf{z}_{t}^e \mid \mathbf{z}_{t-1}^e, \mathbf{z}_{0}^e\right)=\mathcal{N}\left(\mathbf{z}_{t}^e ; \sqrt{1 - \beta_t}\mathbf{z}_{t-1}^e, \beta_t \mathbf{I}\right),
\end{equation}
where $\mathcal{N}\left( \cdot \right)$ is a Gaussian distribution, and $\mathbf{I}$ is an identity matrix. 
Here, rather than completely destruct the original distribution, we get a partially noisy latent code $\mathbf{z}_{T'}^e$, where $0<T'< T$.
It can help to avoid the early steps that move along the high variance axes and focus more on the local target attribute, similar to the discussion about mask estimation.
Another reason is that a smaller step can boost the editing speed.
For clarity, the latent code $\mathbf{z}_{T'}^e$ is denoted as $\mathbf{w}_{T'}^e$ during denoising, while retaining the representation $\mathbf{z}_{T'}^e$ during noise addition.
Different from the mask estimation, we replaced the masked region of $\mathbf{w}_{t}^e$ with $\mathbf{z}_{t}^e$ in each denoising step,
\begin{equation}
\mathbf{w'}_{t}^e =  M\odot\mathbf{w}_{t}^e + \left(1-M\right)\odot\mathbf{z}_{t}^e.
\end{equation}
The full pipeline is simply represented as, 
\begin{equation}
\mathbf{w}_{\rm edit} = D\left(\mathbf{w}_{T'}^e, y_{tgt}, M \right).
\end{equation}
The editing direction discovered in a specific instance can also be generalized to other instances and produce similar attribute editing,
\begin{equation}
\Delta \mathbf{\hat{w}} = \mathbf{w}_{\rm edit} - \mathbf{w}_{\rm e},
\end{equation}
\begin{equation}
I_{\rm edit} = \mathcal{G}\left(\mathbf{w}_{\rm input} + \alpha \Delta \mathbf{\hat{w}}\right).
\end{equation}


\begin{figure*}[t]
	\centering
	\includegraphics[width=\linewidth]{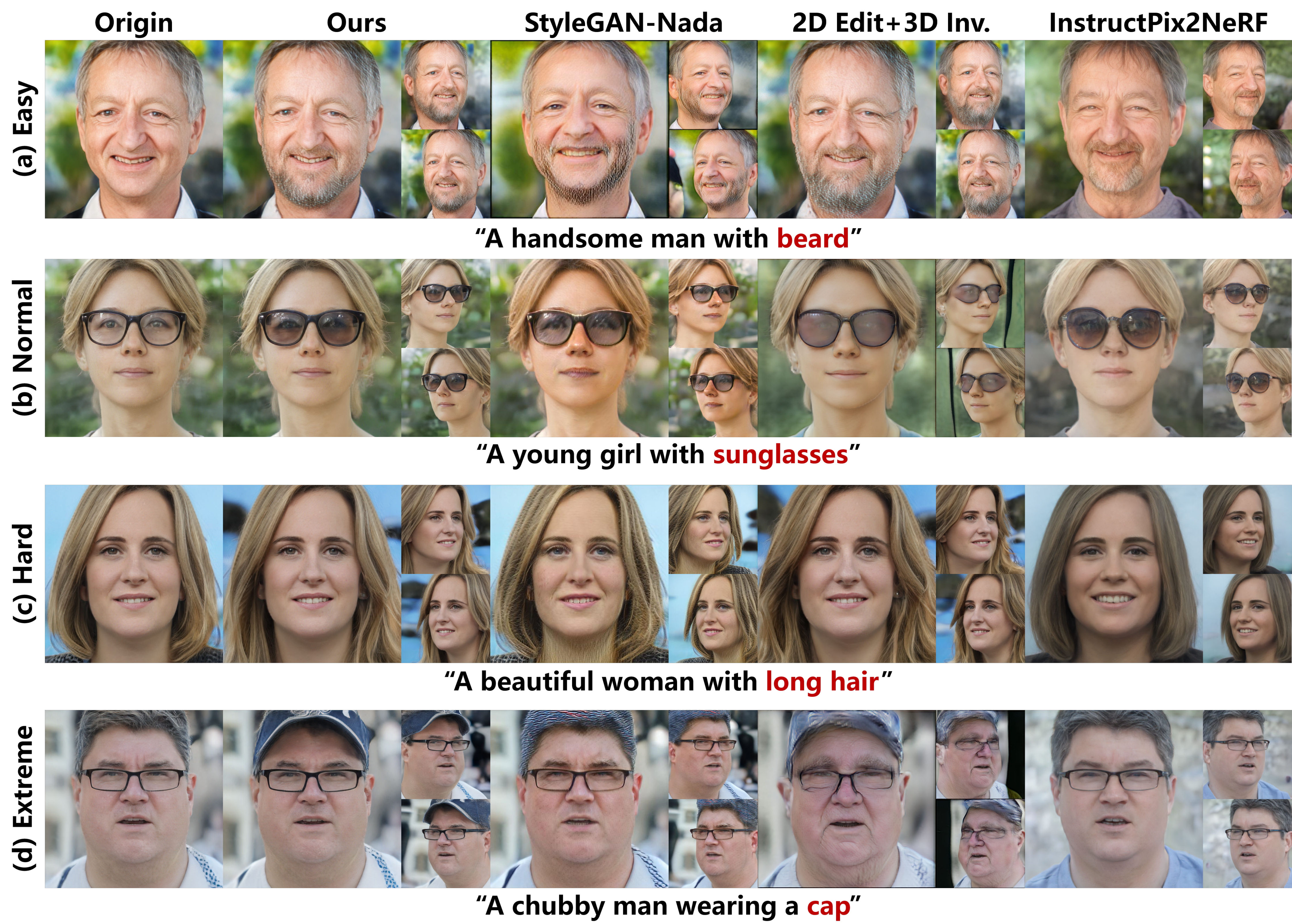}
	\caption{Qualitative comparisons with text-guided face editing methods. (a), (b), (c), and (d) are the cases with empirically different difficulties, \ie easy for beard, normal for sunglasses, hard for hair, and extreme for cap. Our method achieves better results in most cases compared to the other three methods, with higher text consistency and stronger identity preservation.}
	\label{img:comparison}
\end{figure*}

\begin{figure*}[h]
  \centering
  \includegraphics[width=0.9\linewidth]{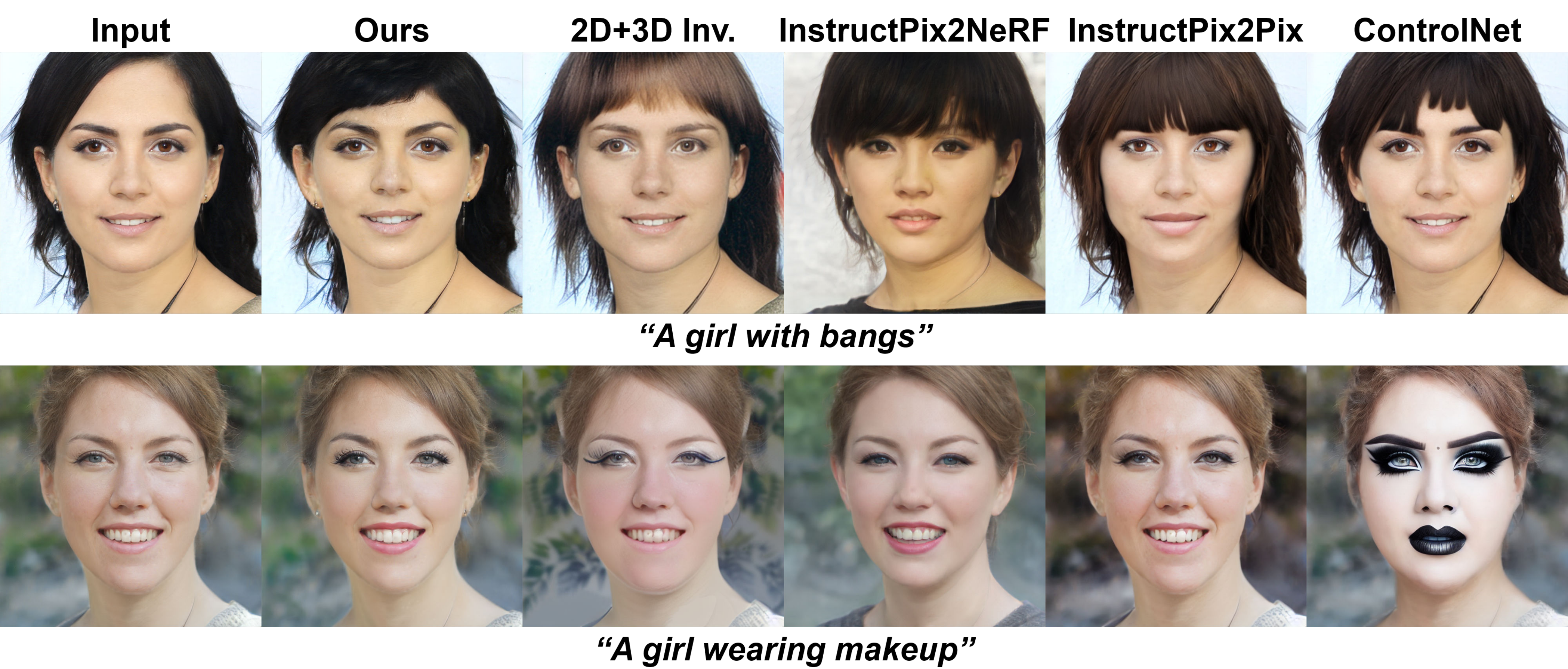}
  \caption{Qualitative comparisons with additional 2D methods. Our method can achieve natural results for both geometry and appearance editing while maintaining good identity consistency.}
  \label{fig:qual_comp_2d}
\end{figure*}

\section{Experiments}

\subsection{{Implementation Details}}
\label{Sec:4-1}
\noindent\textbf{Training Details.} To train our diffusion model, we perform offline synthesis of 1,000,000 images using $\mathbf{w}$ codes obtained from the pre-trained EG3D~\cite{chan2022efficient}.
Subsequently, we extract the features of these images using CLIP~\cite{radford2021learning}.
Similar to previous works~\cite{pinkney2022clip2latent, li20233d}, we incorporate classifier-free guidance and pseudo-text embeddings during training to mitigate overfitting and enhance diversity.
The training procedure for the diffusion model requires approximately 3 days on FFHQ dataset~\cite{karras2019style} and 2 days on AFHQ dataset~\cite{choi2020stargan}. 
The implementation is executed on 1 Nvidia A6000 GPU.

\noindent\textbf{3D GAN Inversion.}
To perform editing on real-world images, we employ both an encoder-based method GOAE~\cite{Yuan_2023_ICCV}, and an optimization method PTI~\cite{roich2021pivotal} for GAN inversion.
We choose the superior outcome from these two inversion methods for each specific case.

\noindent\textbf{Baselines.}
We compare our proposed method with three 3D face editing approaches that utilize text guidance: 2D editing~\cite{hertz2022prompt, brooks2023instructpix2pix} + 3D GAN Inversion~\cite{roich2021pivotal, Yuan_2023_ICCV},
StyleGAN-Nada~\cite{gal2022stylegan}, and InstructPix2NeRF~\cite{li2023instructpix2nerf}. Among these, InstructPix2NeRF is the most closely related to our method, as it combines a diffusion model with a 3D-aware GAN.
StyleGAN-Nada is a fine-tuning-based method that optimizes the parameters of a pre-trained generator supervised by the CLIP directional loss.
In the method of 2D editing + 3D GAN Inversion, we adopt prompt-to-prompt and InstrcutPix2Pix, and select the best results between them for each instance. The methods of 3D GAN inversion align with others.

%
%

\subsection{Qualitative Evaluation}
\label{Sec:4-2}
\noindent\textbf{Appearance Comparison.}
In this section, we conduct comparison experiments between our methods and several text-guided editing methods.
Based on experience, we sequentially present four attributes, each representing a different level of editing difficulty: beard for Easy, sunglasses for Normal, hair for Hard, and cap for Extreme.
First of all, it is significant to clarify the difference between direct manipulation on 3D space and 2D editing + 3D inversion. 
As shown in \cref{img:comparison}, although 2D editing can produce photo-realistic and text-consistent results, the following 3D GAN inversion struggles with some special attributes, \eg, colorful cap and reflective sunglasses.
It results in the loss of details and the ambiguity in appearance, \eg, blurry blue hair instead of a blue cap.
We further compare the results among 3D methods. The combination of StyleGAN-Nada~\cite{gal2022stylegan} with EG3D enables color editing, while it encounters challenges in geometry manipulation. Besides, the overall face color slightly changes.
Although InstructPix2NeRF~\cite{li2023instructpix2nerf} can edit most attributes, it still fails on the most challenging attribute and identity preservation.
In contrast, our method achieves text-consistent and natural editing results under arbitrary text prompts while maintaining identity and disentanglement.
Besides, we also provide a comparison of single-view results with several additional 2D editing methods, including InstrcutPix2Pix~\cite{} and ControlNet~\cite{}. As shown in~\cref{fig:qual_comp_2d}, our method can keep
better ID consistency than others while achieving target editing.

\begin{figure}[t]
	\centering
	\includegraphics[width=\linewidth]{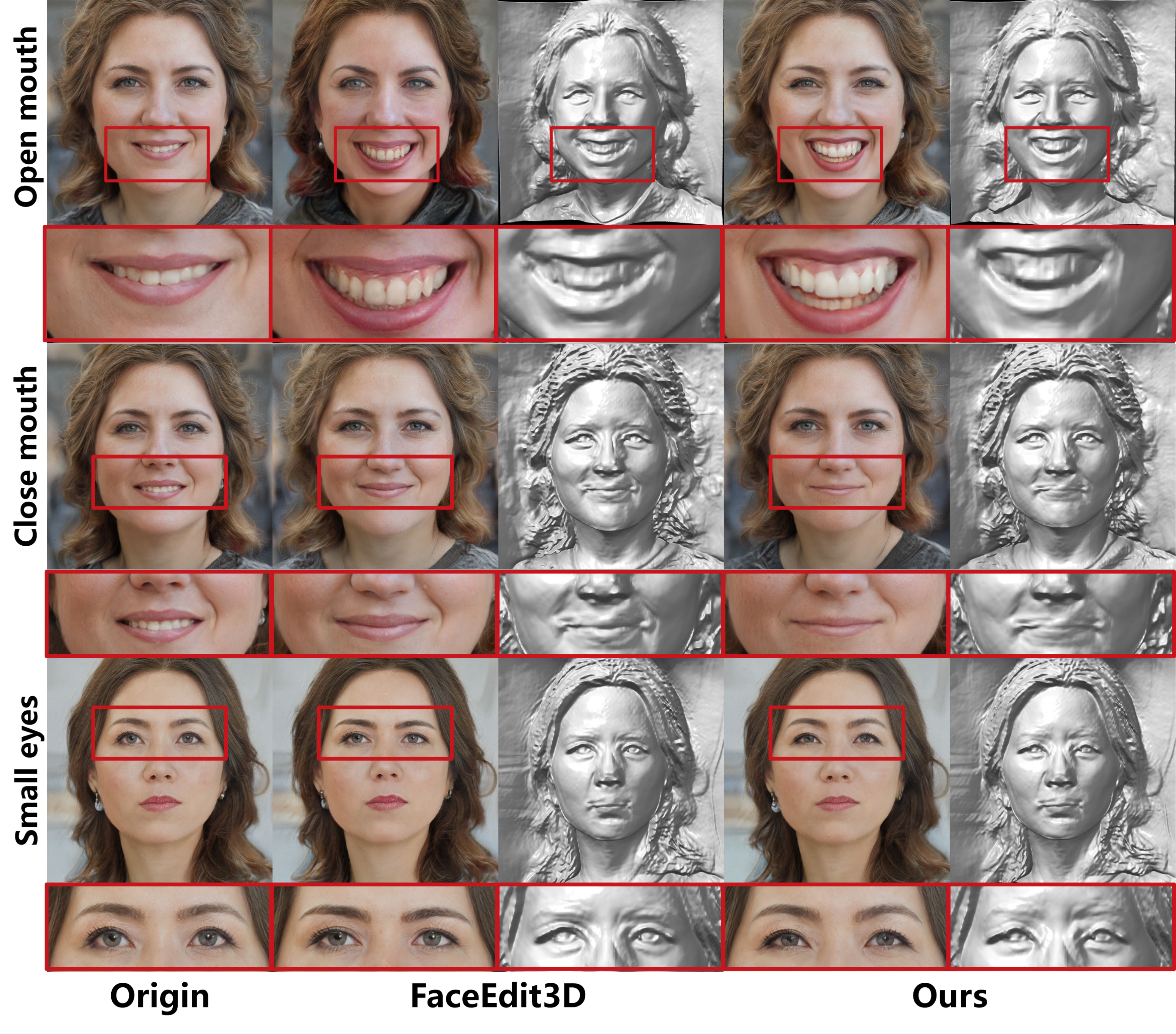}
	\caption{Qualitative comparisons with FaceEdit3D~\cite{Cheng_2024_CVPR} on geometry manipulation.}
	\label{img:geo_comp}
\end{figure}

\begin{figure}[t]
  \centering
  \includegraphics[width=\linewidth]{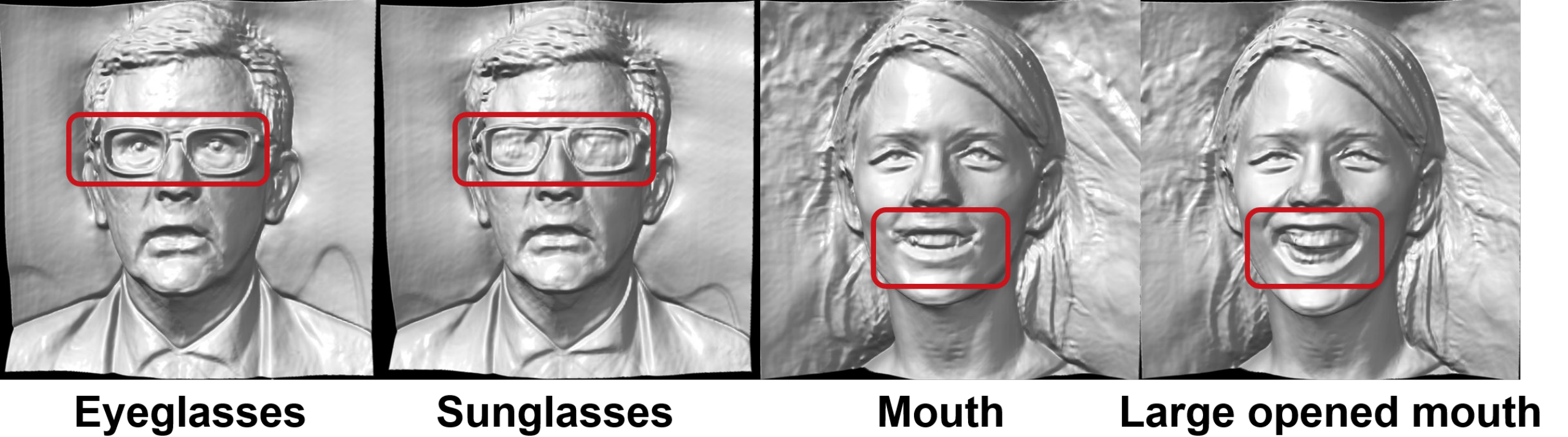}
  \caption{Geometric changes after editing face attributes.}
  \label{geo_changes}
\end{figure}

\noindent\textbf{Geometry Comparison.}
Moreover, an additional comparison of geometric manipulation is made between our work and one of the SOTA 3D dragging methods, FaceEdit3D~\cite{Cheng_2024_CVPR} (CVPR2024). As shown in \cref{img:geo_comp}, we can achieve more natural results in terms of geometry manipulation compared with FaceEdit3D. Our method also maintains better identity consistency, \eg, the case of ``Open mouth''.
It is important to note that FaceEdit3D is a warping-based method that only focuses on geometry manipulation and does not support texture editing. In contrast, our method has the capacity to edit both the geometry and texture with arbitrary attribute descriptions.
To demonstrate the geometric changes, we provide two synthetic cases with before-and-after editing results for visualization in~\cref{geo_changes}. The geometric difference between eyeglasses and sunglasses can provide a better understanding of the 3D role, \ie the consistent change between appearance and geometry.

\begin{table*}[t]
    \centering
    \begin{tabular}{lcccc}
    \toprule
   ~Methods & \makecell{Editing Time $\downarrow$} & \makecell{CLIP Similarity $\uparrow$} & \makecell{ID Similarity $\uparrow$} & $MSE_{o}$ $\downarrow$ ~  \\
     \hline
   ~2D editing~\cite{hertz2022prompt, brooks2023instructpix2pix} + 3D Inv.~\cite{Yuan_2023_ICCV,roich2021pivotal} & 92s & 0.259 & 0.739 & 0.043 ~ \\
   ~StyleGAN-Nada~\cite{gal2022stylegan}~(SIGGRAPH2023) & 107s + 1s & 0.258 & 0.754 & 0.056 ~ \\
   ~InstructPix2NeRF~\cite{li2023instructpix2nerf}~(ICLR2024) & 45s & 0.264 & 0.764 & 0.037 ~ \\
   ~Ours~($\epsilon$ mask) & \textbf{10s} & \textbf{0.273} & \textbf{0.830} & \textbf{0.031} ~ \\
   \hline
   ~Ours~($W$ mask) & 10s & 0.262 & 0.788 & 0.041 ~ \\
   ~Ours~($\mathbf{w}$ swap) & 5s & 0.251 & 0.735 & 0.043 ~ \\
   ~Ours~(w/o mask)& 5s & 0.270 & 0.598 & 0.068 ~ \\
   \bottomrule
    \end{tabular}
        \caption{Quantitative comparison of efficiency and quality. We also conduct an ablation study to analyze the effect of $\epsilon$ mask.
        }
    \label{tab:quantative}
\end{table*}

\begin{table}[t]
    \centering
    \begin{tabular}{lcc}
    \toprule
   ~Methods & 3D CLIP Score $\uparrow$ & Multi-view ID Sim. $\uparrow$~  \\
     \hline
   ~2D + 3D Inv. & 0.252 & 0.517 ~ \\
   ~StyleGAN-Nada & 0.238 & 0.533 ~ \\
   ~InstructPix2NeRF & 0.249 & 0.468 ~ \\
   ~Ours & \textbf{0.255} & \textbf{0.560} ~ \\
   \bottomrule
    \end{tabular}
        \caption{Quantitative comparison of multi-view results.
        }
    \vspace{-0.3cm}
    \label{tab:multi-view}
\end{table}

\subsection{Quantitative Evaluation}
\label{Sec:4-3}

In quantitative experiments, we choose 10 editing directions, each of which samples 5 identities, to evaluate the efficiency, text consistency, ID consistency, and attribute disentanglement.
As shown in \cref{tab:quantative}, we adopt the editing time as the metric for the efficiency measurement. 
For real-world image editing, we have subtracted the time spent on the GAN inversion for all methods. 
As shown in \cref{tab:quantative}, our method achieves the fastest editing time within 10 seconds, while StyleGAN-Nada spends a long time on generator fine-tuning for each input text. 
Although 2D editing + 3D inversion and InstructPix2NeRF are both diffusion-based methods like ours, they are still time-consuming, as they require an additional diffusion inversion for 2D image editing at the initial step.
Furthermore, we compare CLIP similarity and ID similarity to measure the effectiveness of text-guided editing.
The highest scores achieved by our methods in both metrics demonstrate our ability to balance text consistency and ID preservation, which are the most essential aspects of the text-guided editing task.
We further adopt the pixel-wise MSE outside the target region to evaluate the attribute disentanglement. Our method also achieves the lowest error in irrelevant regions.
The best performance across all four evaluation metrics proves the superiority of our methods.

As it is hard to directly evaluate the geometric results during editing, here we additionally provide the 3D quantitative results under multi-view metrics. The metrics are similar to those in single-view quantitative results, except that we render images from two random side views and
measure the similarity of those two views.
As shown in~\cref{tab:multi-view}, our method also outperforms others in multi-view
evaluations, demonstrating the superiority of our methods in 3D space.

\begin{figure}[t]
	\centering
	\includegraphics[width=\linewidth]{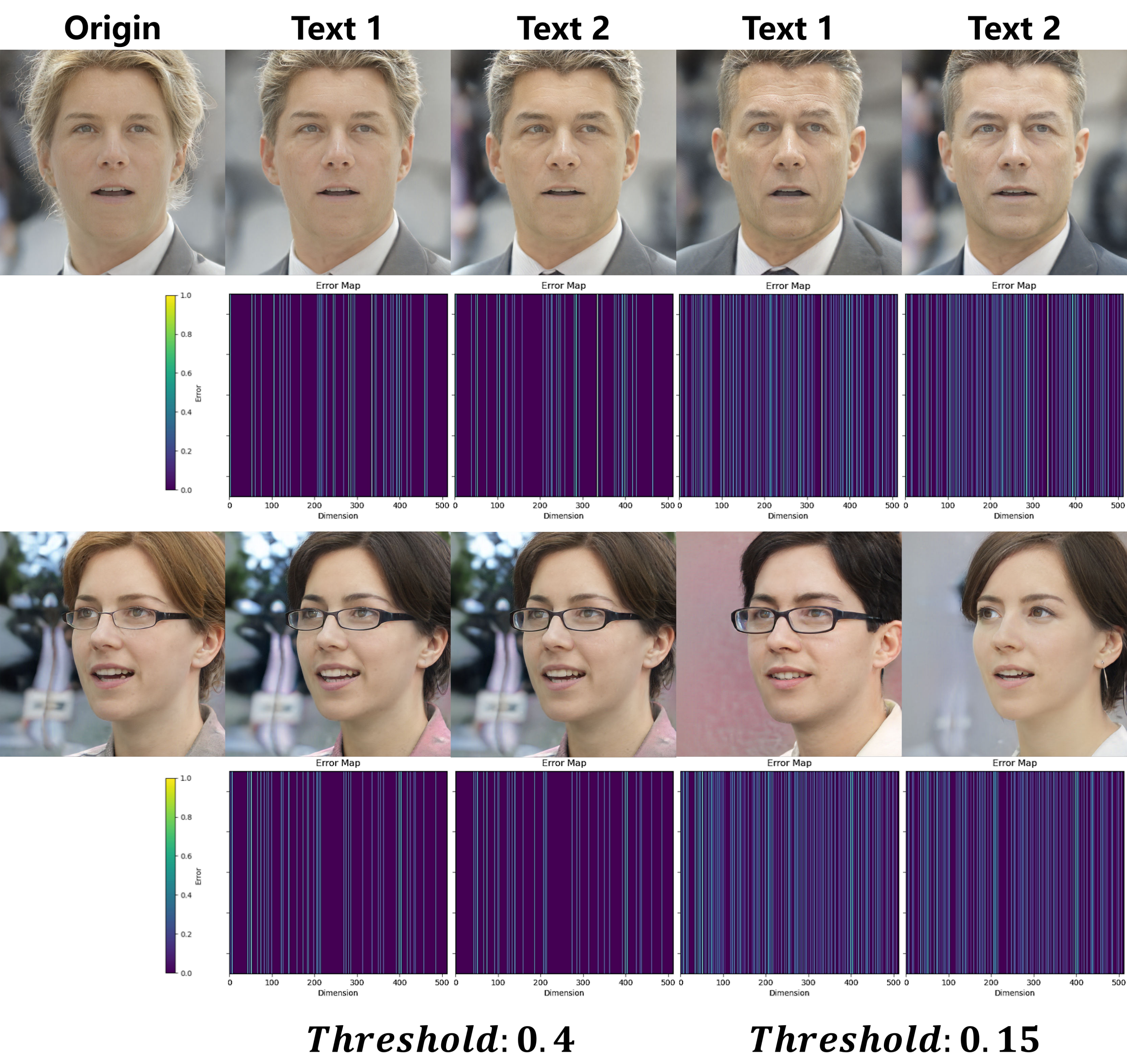}
	\caption{Mask estimation from different text pairs. The paired texts 1 in the first example are \textit{``a person with short hair''} and \textit{``a person with long hair''}. The paired texts 2 in the first example are \textit{``a long-haired man''} and \textit{``a short-haired man''}. The paired texts 1 in the second example are \textit{``a smart girl with blonde hair''} and \textit{``a smart girl with black hair''}. The paired texts 2 in the second example are \textit{``a blond person''} and \textit{``a black-haired person''}. Please zoom in for detailed observation on mask distribution.
 }
	\label{img:disentangle}
\end{figure}

\begin{figure}[t]
	\centering
	\includegraphics[width=\linewidth]{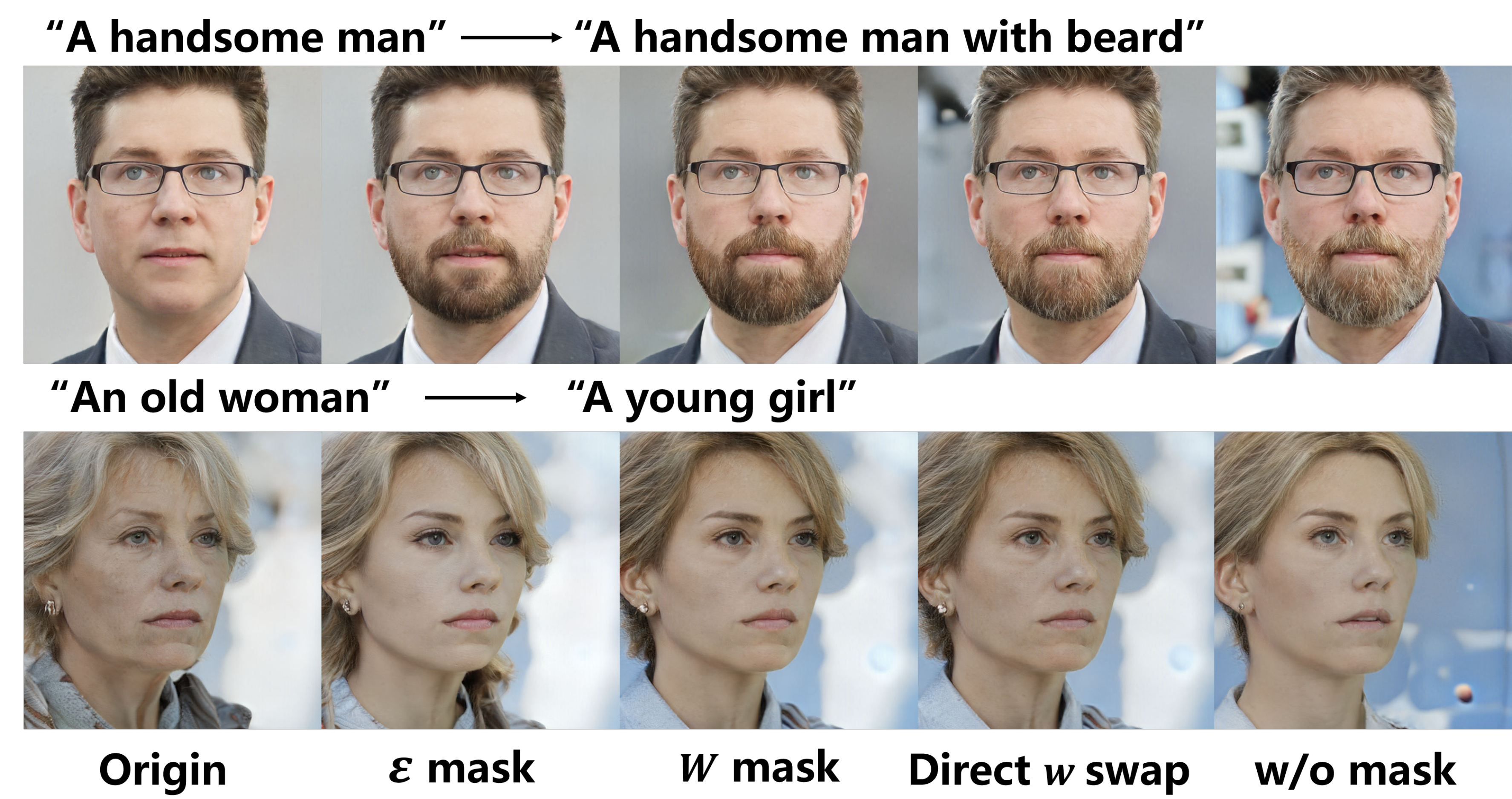}
	\caption{The ablation study of the latent mask.}
	\label{img:ablation}
 \vspace{-0.1cm}
\end{figure}

\subsection{Ablation Study}
\label{Sec:4-4}
\noindent\textbf{Text-guided Mask Estimation.}
To investigate the robustness of our text-guided mask estimation, we conduct an ablation study to demonstrate the variation of an attribute's mask on latent code when using different text pairs with similar meanings.
For example in \cref{img:disentangle}, we use \textit{``a person with short hair''} and \textit{``a short-haired man''} as different target text prompts to identify the latent mask for editing hair length manipulation.
We visually represent the latent masks as error maps together with their corresponding edited faces.
As illustrated in \cref{img:disentangle}, they exhibit nearly the same latent masks and editing results with different target prompts within an appropriate threshold.
It proves that the relative direction of paired texts is stable and effective in estimating the mask.
The experiment also reveals an interesting phenomenon that the results in low threshold include obvious changes in age besides hair length.
This can be attributed to our generative mapper's tendency to associate certain attributes according to data distribution, \eg, sparse hair and old age, thick hair and youthfulness.
The experiment shows that our estimated mask serves as a filter, eliminating these associated attributes while preserving the primary attribute.
A similar trend is observed in the hair color editing. 

\noindent\textbf{Mask-guided Direction Discovery.}
We further conduct an ablation study on the effect of different ways to apply the mask.
Compared with mask-guided ways, results without mask guidance expose inconsistencies in identity and other irrelevant attributes, \eg, hairstyles.
Directly applying the first-step $\mathbf{w}$ direction to the masked region proves to be effective, benefiting from the accurate estimation of the mask.
However, to precisely filter out text-irrelevant attributes, the magnitude of change in the primary attribute is constrained, as shown in the \textit{``Age''} case in \cref{img:ablation}.
Despite the scalability of the editing direction, a large scale leads to the interference of other attributes due to the incomplete disentanglement of this direction, as shown in the \textit{``Beard''} case in \cref{img:ablation}.
Finally, we compare the masks estimated from the paired predicted noises with those from the paired latent codes.
The difference between paired latent codes can be understood as the cumulative bias of full-step noises.
It contains more global changes, as the initial steps primarily contribute to global synthesis, while the subsequent steps primarily contribute to local generation.
Compared to masks from the paired latent codes, masks from the last few noises show better disentanglement.
We additionally conduct a quantitative comparison in \cref{tab:quantative} to support the analysis.



\subsection{Applications}
\label{Sec:4-5}

\begin{figure}[t]
    \centering
    \includegraphics[width=\linewidth]{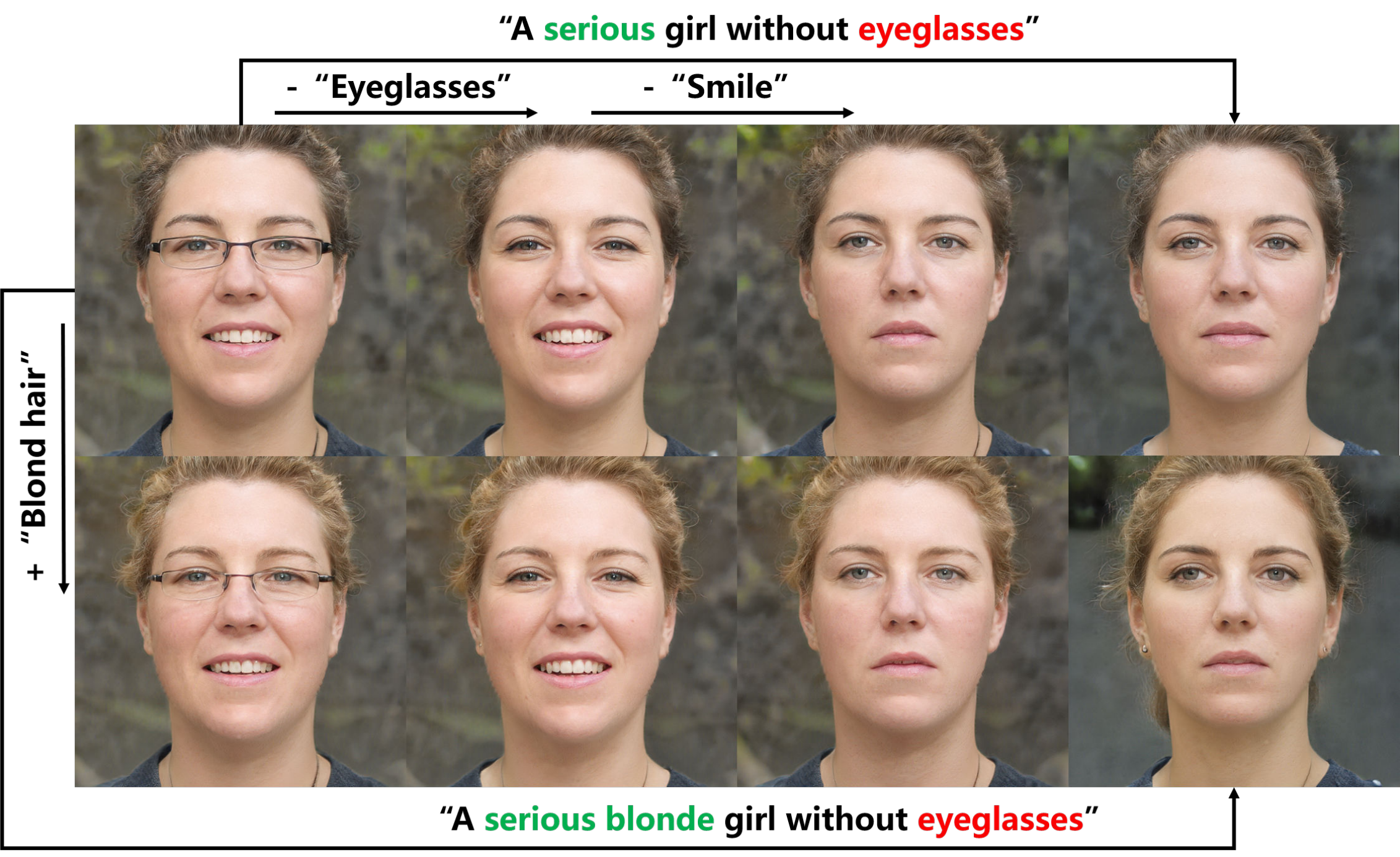}
    \caption{The illustration of continuous single-attribute editing and simultaneous multi-attribute editing.}
    \label{img:multy}
\end{figure}

\begin{figure}[t]
  \centering
  \includegraphics[width=\linewidth]{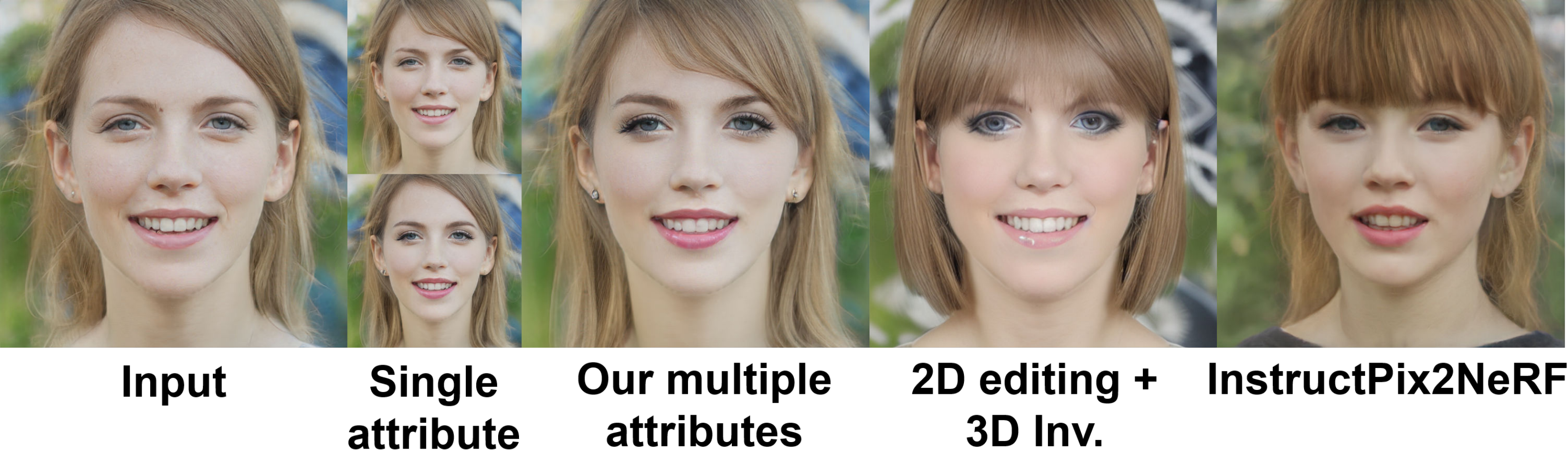}
  \caption{Comparison of multi-attribute editing.}
  \label{fig:multi_comp}
  \vspace{-0.1cm}
\end{figure}

\begin{figure}[t]
    \centering
    \includegraphics[width=\linewidth]{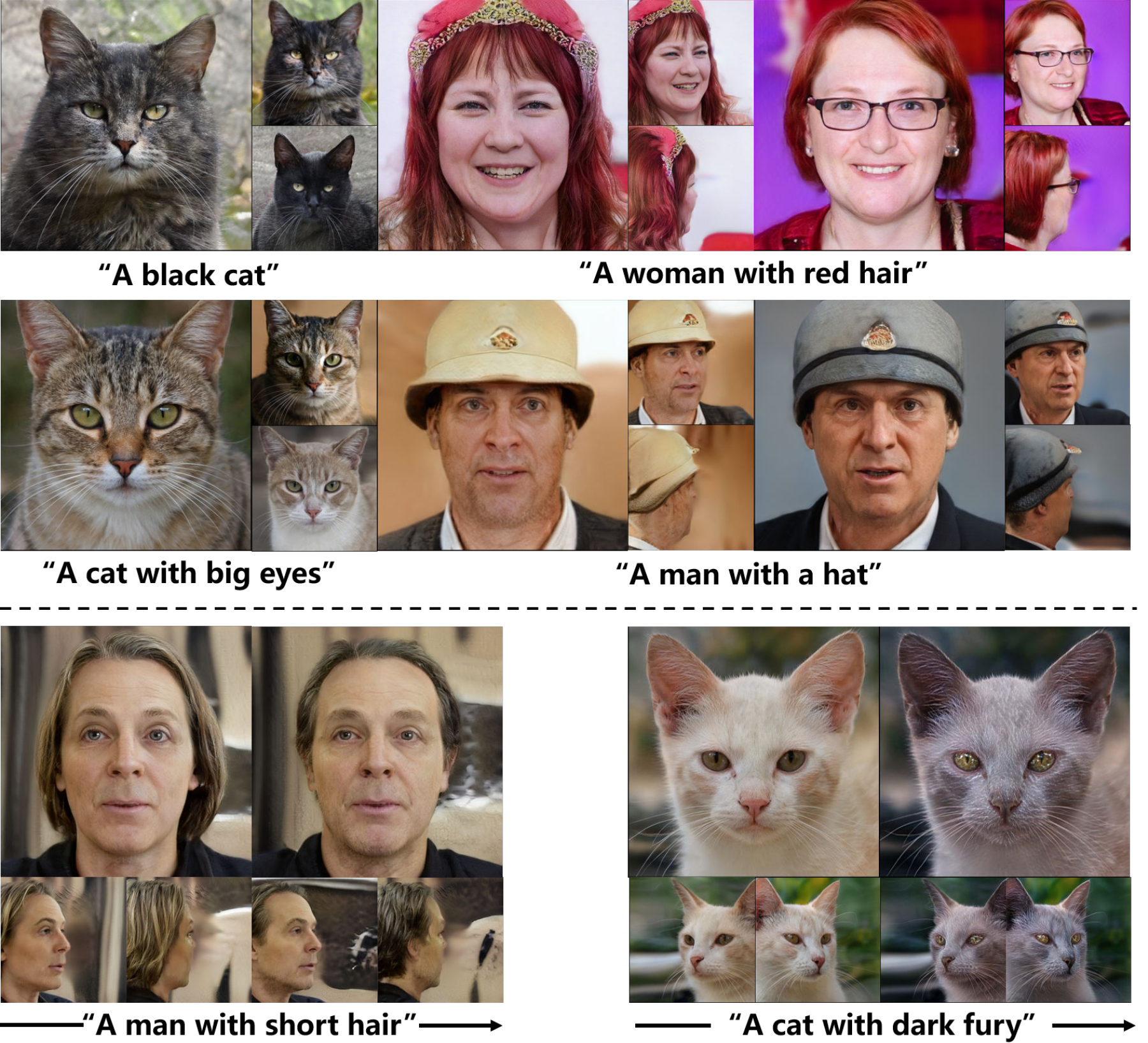}
    \caption{Generalization to cat faces and full heads.
    }
    \vspace{-0.1cm}
    \label{img:general}
\end{figure}

\noindent\textbf{Multiple-attributes Editing.}
Our method supports both continuous single-attribute editing and simultaneous multiple-attribute editing.
As shown in \cref{img:multy}, we gradually edit the original face with \textit{``without eyeglasses''}, \textit{``serious''}, and \textit{``blond hair''}. The results demonstrate our capacity for ID preservation and attribute disentanglement during continuous editing.
We conduct a comparison using a multi-concept prompt to directly edit multiple attributes, \ie \textit{``A serious girl without eyeglasses''} and \textit{``A serious blonde girl without eyeglasses''}.
Despite minimal entanglement, the natural results prove that our method can support the simultaneous editing of three attribute targets in a text prompt.
We also provide a comparison of multi-attribute editing with other 3D methods. As shown in~\cref{fig:multi_comp}, our method better keeps the identity during editing multiple attributes than other methods.

\noindent\textbf{Generalization to Other Generators.}
Our method can be generalized to other generators, \eg, cat faces or full heads.
As shown in \cref{img:general}, the generated samples exhibit diversity and photo-realism, while edited samples are identity-consistent.
The experiment demonstrates that our architecture can generate and manipulate full heads and cat faces on other pre-trained generators, consistent with the user-provided text guidance.


\section{Conclusion}
In this paper, we propose Face Clan, a fast and generalized method for text-guided 3D face editing. Compared to previous text-guided 3D face editing, our method does not require optimization but can handle arbitrary attribute descriptions. Leveraging a pre-trained 3D-aware GAN, we construct a self-supervised diffusion model to align the text distribution with the latent manifolds of the GAN. It serves as a mapping network that projects noises to semantic meaningful latent codes, while its classifier capability empowers us to reveal the editing direction in the latent space. To enhance the disentanglement, we propose estimating a semantic mask in latent space based on pairs of opposite descriptions. The mask significantly preserves the original identity and irrelevant attributes. Experiments demonstrate the efficiency, generalization, and effectiveness of our methods in both synthetic and real-world face editing.
As we conclude our exploration into text-guided image editing, our study well incorporates textual guidance to image manipulation, advancing the frontiers of multimedia innovation and content creation.

\noindent{\bf Limitations and Future Work.}
Despite natural and text-consistent results, our method also encounters several limitations.
Due to the curved trajectory of DDIM or DDPM sampling, the mask estimation needs to measure the difference of predicted noise in the last few steps which is inefficient. Besides, it leads to up to 5 seconds during denoising, still a long time for users. To solve the problem, we are exploring the possibility of introducing Rectified Flow~\cite{liu2022flow, lipman2022flow} whose trajectory is a straight line. The straight trajectory may improve the editing speed and stability.
Besides, it is hard for our method to manipulate some corner attributes, \eg, wearing a headset and eyepatch, due to the limited capacity of pre-trained 3D-aware GANs.

{
\bibliographystyle{IEEEtran}
\bibliography{main}
}

\newpage

 




\vfill

\end{document}